\pdfoutput=1
\documentclass[runningheads]{llncs}

\usepackage{eccv}

\usepackage{eccvabbrv}
\usepackage{graphicx}
\usepackage{booktabs}
\usepackage{capt-of}
\usepackage[accsupp]{axessibility}
\usepackage{amsmath}
\usepackage{amssymb}
\usepackage{hyperref}

\graphicspath{{figures/}}

\newcommand{\name}{PeakPatch}

\begin{document}

\title{What CLIP Knows but Cannot Say: Recovering Negation from Frozen Intermediate Features}

\titlerunning{What CLIP Knows but Cannot Say}

\author{Chen-Yi Lu \and
Yueh-Shao Chen \and Somali Chaterji}

\authorrunning{Chen-Yi Lu et al.}

\institute{Purdue University, West Lafayette, IN 47907, USA}

\maketitle

\begin{figure}[t]
  \centering
  \includegraphics[width=0.75\linewidth]{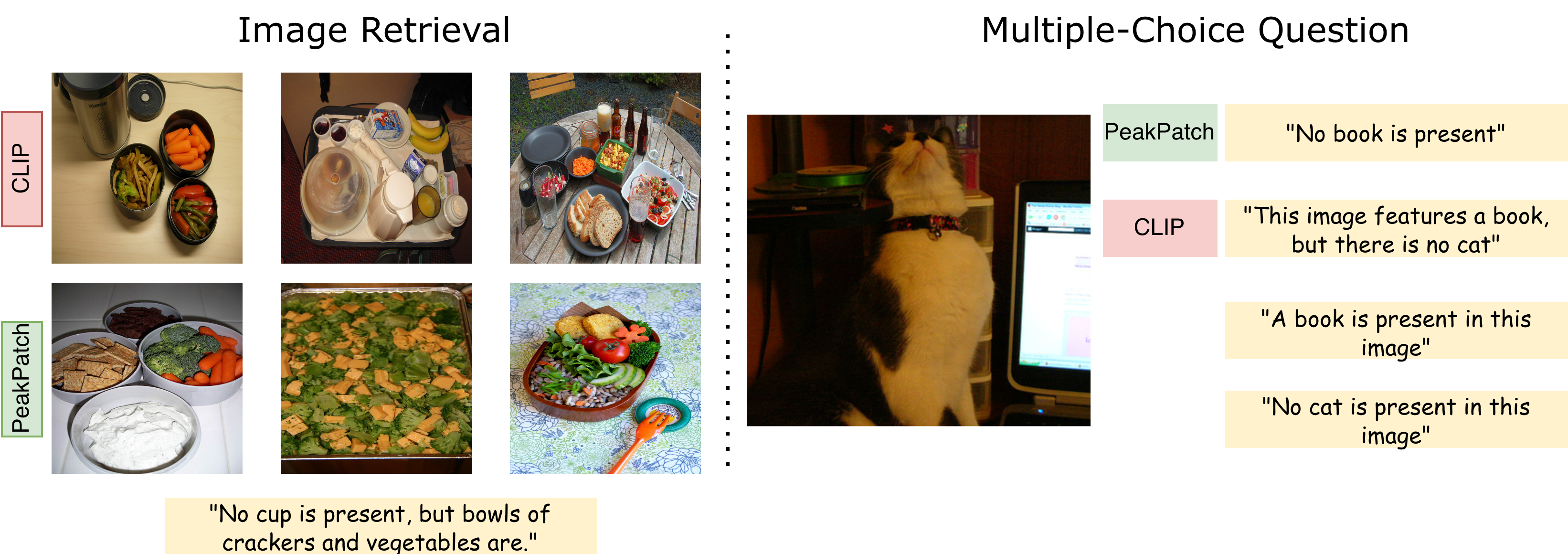}
  \caption{\textbf{CLIP's negation blindness on two NegBench protocols.} (\emph{Left}) Retrieval: CLIP retrieves images containing the negated object (cup), treating ``No cup is present'' as similar to ``cup present.'' \name{} correctly retrieves cup-free images. (\emph{Right}) MCQ: given a cat image with no book, CLIP selects a caption mentioning both nouns while ignoring negation; \name{} selects the correct answer (``No book is visible in this image.'').}
  \label{fig:teaser}
\end{figure}

\begin{abstract}
Contrastive vision-language models such as CLIP map semantically opposite phrases (\eg, ``a dog'' \vs ``not a dog'') to nearly identical embeddings, rendering them insensitive to negation.
We attribute this failure to a phenomenon we call \emph{Representational Collapse}: by tracking compositional divergence and visual alignment across the CLIP text encoder, we show that middle layers build compositional syntax, but the final layers collapse this structure as visual alignment rises, producing a syntax-blind final representation.
To recover the lost negation signal without altering pretrained weights, we propose \name, a lightweight post-hoc correction system that intercepts the encoder at its \textit{compositional peak}
while keeping CLIP fully frozen.
An Embedding Correction Network (ECN) uses cross-attention to extract a negation-specific signal from the peak layer, anchored to a stable baseline, and predicts a deviation vector that re-injects the lost syntax into the final-layer embedding space.
A complementary Score Correction Network (SCN) predicts bounded scalar score offsets for discriminative tasks.
Both modules are trained jointly end-to-end while all CLIP parameters remain frozen, adding only 5.2M parameters (3.5\% of the backbone) and preserving the standard cosine similarity interface.
On NegBench, \name{} achieves 74.3\% on COCO MCQ (+35.1 over CLIP, +17.8 over the best encoder fine-tuning method) and 65.5\% on VOC MCQ, while outperforming all fine-tuning baselines on fully out-of-distribution negation retrieval despite training only 3.5\% of the parameters.
The corrected embeddings also transfer to text-to-image generation (+18.4 negation score) and generalize across ViT-B/32, ViT-L/14, and SigLIP backbones.

\noindent Project page: \url{https://stevencylu.github.io/PeakPatch/}
\end{abstract}

\section{Introduction}
\label{sec:intro}

Vision-language models (VLMs) such as CLIP~\cite{radford2021clip} learn a shared embedding space where images and text can be directly compared via cosine similarity.
This simple yet powerful formulation has enabled a broad range of applications, including zero-shot classification~\cite{radford2021clip,zhou2022coop}, cross-modal retrieval~\cite{yuksekgonul2023negclip,luo2022clip4clip}, text-to-image generation~\cite{rombach2022ldm,ramesh2022dalle2}, open-vocabulary detection~\cite{li2022glip,kamath2021mdetr,yao2022detclip,liu2024groundingdino}, and semantic segmentation~\cite{luddecke2022clipseg,xu2022groupvit}.
Subsequent models such as ALIGN~\cite{jia2021align}, OpenCLIP~\cite{cherti2023openclip}, SigLIP~\cite{zhai2023siglip}, and EVA-CLIP~\cite{sun2023evaclip} have scaled this paradigm to larger datasets and architectures.

Despite this progress, these models share a fundamental limitation: they cannot distinguish negated descriptions from affirmative ones.
Semantically opposite phrases (\eg, ``a dog'' \vs ``not a dog'') are mapped to nearly identical embeddings, making the model effectively blind to negation.
As \cref{fig:teaser} illustrates, this causes CLIP to retrieve images containing the negated object and to select wrong captions in multiple-choice settings.
On the NegBench VOC MCQ task~\cite{alhamoud2025negbench}, this manifests as a universal \emph{affirmation bias}: 81\% accuracy on affirmative captions collapses to just 3\% on negated ones, worse than random chance.
This limitation has practical consequences: applications routinely require negation, from a radiologist searching for ``bilateral consolidation with no evidence of pneumonia'' to a safety inspector querying ``construction sites with no barriers'' to content moderation prompts such as ``no weapon.''

Why does this happen, and can it be fixed?
Kang~\etal~\cite{kang2025clip} prove that the problem is fundamental: no CLIP-like joint embedding space can correctly handle even \emph{any two} of basic semantics, attribute binding, spatial relations, and negation, as the geometry is overconstrained.
Fine-tuning methods~\cite{yuksekgonul2023negclip,singh2025conclip,garg2025negationclip} improve negation accuracy without large drops in zero-shot classification, but they remain subject to this geometric ceiling---they push negation into a space that provably cannot fully accommodate it.
Moreover, fine-tuning permanently alters the pretrained weights, requiring every downstream system that builds on frozen CLIP features (\eg, LLaVA~\cite{liu2023llava}, BLIP-2~\cite{li2023blip2}, text-to-image generators~\cite{rombach2022ldm}) to be re-validated or re-adapted.
We therefore ask: \emph{Can we correct CLIP's negation blindness without modifying the encoder or relying on an external model?}

\begin{figure}[t]
  \centering
  \includegraphics[width=0.65\linewidth]{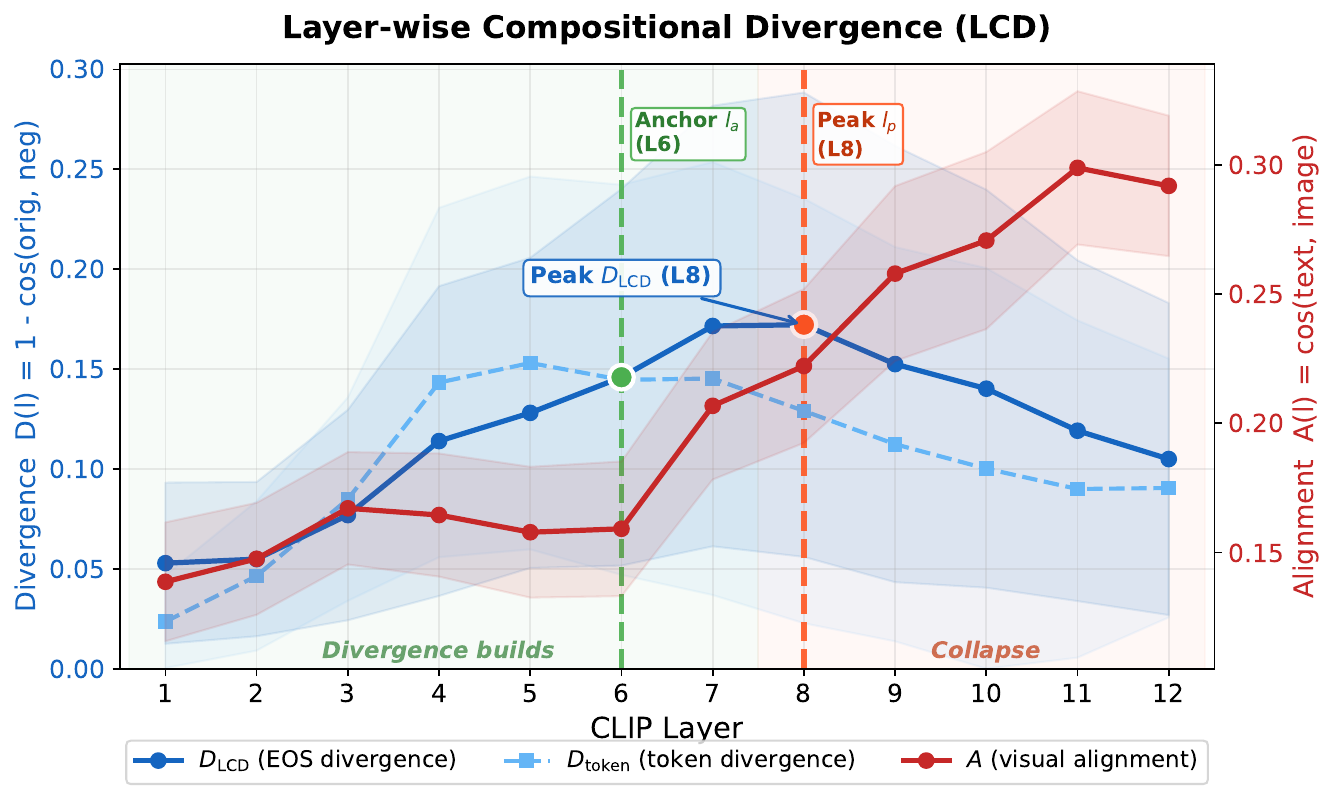}
  \caption{\textbf{Representational Collapse in CLIP's text encoder.} 
  We track two layer-wise signals: compositional divergence (LCD; blue, left axis), defined as $D(\ell)=1-\cos(\text{orig},\text{neg})$, and visual alignment (red, right axis), defined as $A(\ell)=\cos(\text{text},\text{image})$.
  Divergence peaks at Layer~$l_p$ but drops sharply as alignment rises, collapsing into a syntax-blind representation. \name{} extracts negation features at $l_p$ (compositional peak) and anchors them to $l_a$ (L6) to correct the collapsed final-layer output.}
  \label{fig:motivation}
\end{figure}

To answer this, we probe the frozen CLIP text encoder, layer by layer.
We introduce the Layer-wise Compositional Divergence (LCD) metric, which tracks how well each layer separates negated captions from their affirmative counterparts (\cref{fig:motivation}).
Jointly measuring this compositional divergence alongside per-layer visual alignment to the paired image reveals a clear trajectory.
LCD rises through early and middle layers as the encoder builds compositional structure, then peaks at a \emph{compositional peak} layer $l_p$.
Visual alignment, meanwhile, remains low through an anchor layer $l_a$ and then rises sharply.
Beyond $l_p$, visual alignment dominates while LCD drops. The final layers collapse the compositional structure into a syntax-blind representation optimized for cross-modal matching.
We call this phenomenon \emph{Representational Collapse}.

Representational Collapse suggests a simple strategy: extract negation information before the encoder discards it, then inject it back into the standard final embedding interface.
%
Since the final layer is strongly visually aligned but syntax-blind, while the compositional peak is syntax-aware but not yet visually aligned, neither representation suffices alone.
Based on this insight, we propose \name{}, a lightweight post-hoc correction system that intercepts the frozen CLIP encoder at its compositional peak and corrects the collapsed final-layer output.
An Embedding Correction Network (ECN) uses cross-attention to extract a negation-specific signal from Layer~$l_p$, anchors it to a stable baseline at Layer~$l_a$, and predicts a deviation vector that injects the lost syntax into the final-layer embedding space.
A Score Correction Network (SCN) contrasts the peak-layer representation with the ECN-corrected embedding and the image to predict bounded scalar corrections for discriminative tasks.
Both modules are trained jointly end-to-end while all CLIP parameters remain frozen, adding 5.2M parameters (3.5\% of the backbone) and preserving the standard cosine similarity interface.
Overall, we correct CLIP's negation blindness entirely post-hoc, with a frozen encoder and no external inference-time model.
%
Our contributions are as follows:
\begin{enumerate}
  \item We introduce the Layer-wise Compositional Divergence (LCD) metric and use it to identify \emph{Representational Collapse} in CLIP's text encoder: negation separability peaks at an intermediate layer and then degrades as visual alignment increases, revealing that the encoder actively discards compositional structure in its final layers. 
  \item Motivated by this finding, we propose \name{}, a lightweight post-hoc correction system that keeps all of CLIP frozen and requires no external model. An Embedding Correction Network (ECN) extracts negation signals from the compositional peak via cross-attention and corrects the final text embedding, while a Score Correction Network (SCN) applies bounded corrections to the similarity score.
  \item \name{} achieves 74.3\% on COCO MCQ (+35.1 over CLIP, +17.8 over the best encoder fine-tuning method) with only 5.2M parameters (3.5\% of the backbone). In the fully OOD retrieval setting, it outperforms all fine-tuning baselines despite modifying only the text side, and its corrected embeddings transfer to text-to-image generation (+18.4 negation score). 
\end{enumerate}

\section{Related Work}
\label{sec:related_work}

\subsection{Post-Hoc Adaptation of Vision-Language Models}

Contrastive VLMs such as CLIP~\cite{radford2021clip}, ALIGN~\cite{jia2021align}, OpenCLIP~\cite{cherti2023openclip}, SigLIP~\cite{zhai2023siglip}, and FLAVA~\cite{singh2022flava} enable strong zero-shot transfer, but adapting them without damaging pretrained representations remains an open challenge.
Existing approaches operate at three levels of the inference pipeline.
\emph{Prompt learning}~\cite{zhou2022coop,zhou2022cocoop,khattak2023maple,jia2022vpt} optimizes continuous input tokens while keeping weights frozen, yielding gains on few-shot classification but leaving the similarity function unchanged.
\emph{Feature adapters}~\cite{gao2024clipadapter,zhang2022tipadapter,udandarao2023susx} apply residual or cache-based transformations to output embeddings, demonstrating that small feature-space perturbations can produce large task-level improvements.
\emph{Score- and classifier-level corrections}~\cite{yu2023taskres,huang2024lp,zhu2023ape,tang2024amu} modify classifier weights or logits to steer predictions without retraining the backbone.
Complementing these, WiSE-FT~\cite{wortsman2022wiseft} shows that interpolating fine-tuned and zero-shot \emph{model weights} preserves robustness, and parameter-efficient methods such as low-rank adaptation~\cite{hu2022lora} have become standard for updating large models with minimal overhead.
Notably, generative VLMs that use CLIP as a frozen visual encoder~\cite{li2021albef,alayrac2022flamingo,liu2023llava,li2023blip2,dai2023instructblip} inherit any compositional failure in CLIP's representations, making upstream corrections especially consequential.

Despite their diversity, these methods share a common assumption: the embedding geometry is fundamentally sound and merely needs task-specific steering.
\name{} draws on the idea of residual corrections~\cite{yu2023taskres,gao2024clipadapter} but addresses a qualitatively different problem: the text encoder itself produces systematically wrong representations for negated inputs, requiring correction at the representation level before any downstream scoring can be meaningful.

\subsection{Negation Understanding in Vision-Language Models}

Among the compositional failures exposed by recent benchmarks~\cite{thrush2022winoground,ma2023crepe,hsieh2023sugarcrepe,doveh2023dac,huang2024structureclip,zhao2022vlchecklist}, negation stands out as uniquely severe.
While negation has been studied extensively in natural language understanding~\cite{hossain2022negation}, its impact on VLMs has only recently been quantified: NegBench~\cite{alhamoud2025negbench} reveals that CLIP’s 81\% affirmative accuracy collapses to 3\% on negated captions, and CC-Neg~\cite{singh2025conclip} confirms this failure at scale with 228K pairs.
The root cause appears structural rather than data-driven: Kang~\etal~\cite{kang2025clip} prove a geometric impossibility theorem showing that no single embedding space can correctly handle even any two of basic semantics, attribute binding, spatial relations, and negation without overconstraining the geometry,
while Quantmeyer~\etal~\cite{quantmeyer2024clip} provide mechanistic evidence that negation processing is distributed across 
layers 
and concentrated in a small fraction of negator-selective attention heads (8\%).

Existing fixes fall into two camps, each with clear tradeoffs.
\emph{Encoder fine-tuning} methods~\cite{yuksekgonul2023negclip,singh2025conclip,garg2025negationclip} retrain the text encoder on negation-aware data.
This improves negation understanding but risks degrading the broad representations that make CLIP useful, precisely because it forces negation into an embedding space that provably cannot accommodate it~\cite{kang2025clip,wortsman2022wiseft}.
\emph{Inference-time methods} avoid modifying CLIP but introduce other constraints: DCSM~\cite{kang2025clip} replaces cosine similarity with token-to-patch CNN scoring, breaking the standard CLIP interface and preventing use in downstream pipelines that rely on embeddings.
Two concurrent methods fall outside these camps but introduce their own constraints: SpaceVLM~\cite{ranjbar2025spacevlm} and Aggarwal~\etal~\cite{aggarwal2026seeing} explicitly extract the negated concept at inference time, the former through an LLM parser and the latter through a rule-based parser, making them not directly comparable to methods that operate on the raw caption without external parsing. We provide detailed comparisons in \cref{sec:parser_comparison}.
\subsection{Interpreting CLIP’s Internal Representations}

A growing body of work probes what CLIP’s layers encode.
Gandelsman~\etal~\cite{gandelsman2024interpreting} project each ViT layer’s output through CLIP’s final projection head, decomposing the image representation into per-layer contributions.
On the text side, CLIP behaves largely as a bag of words~\cite{yuksekgonul2023negclip}, and Quantmeyer~\etal~\cite{quantmeyer2024clip} use causal tracing to establish that negation awareness \emph{does} exist in intermediate layers but spans only 8\% of attention heads and does not survive to the final representation.
These studies reveal \emph{what} CLIP fails at and \emph{where} information resides, but stop short of quantifying the layer-wise dynamics or translating diagnostics into a correction strategy.

Our LCD analysis (\cref{sec:lcd}) builds on these findings.
Adopting the projection methodology of Gandelsman~\etal~\cite{gandelsman2024interpreting} on the text encoder side, we track the competition between compositional divergence and visual alignment across every layer, revealing a consistent collapse of negation separability in the final layers---a phenomenon we term \emph{Representational Collapse}.
This directly motivates \name{}: lightweight correction modules that read negation signals from intermediate layers where prior mechanistic analyses localize them~\cite{quantmeyer2024clip}, sidestep the impossibility theorem~\cite{kang2025clip} by operating \emph{outside} the joint embedding space, and apply corrections at both the embedding and score levels---all while keeping CLIP frozen.

\section{Method}
\label{sec:method}

\subsection{Preliminaries}
\label{sec:preliminaries}

CLIP encodes images and text into a shared $d$-dimensional space.
Given an image $I$ and text $T$, their similarity is computed as:
\begin{equation}
  s(I, T) = \cos(\mathbf{f}_I, \mathbf{f}_T) = \frac{\mathbf{f}_I^\top \mathbf{f}_T}{\|\mathbf{f}_I\| \, \|\mathbf{f}_T\|},
  \label{eq:cosine}
\end{equation}
where $\mathbf{f}_I = E_I(I) \in \mathbb{R}^d$ and $\mathbf{f}_T = E_T(T) \in \mathbb{R}^d$ are the image and text embeddings, respectively.
The text encoder is an $L$-layer transformer~\cite{vaswani2017attention}; each layer applies multi-head self-attention and a feed-forward sub-network, producing hidden states $\mathbf{H}^l \in \mathbb{R}^{N \times d}$ at layer $l$, where $N$ is the sequence length.
The final text embedding is obtained by projecting the \texttt{[EOS]} token of the last layer through a layer norm and linear head into the shared space: $\mathbf{f}_T = W_T\,\mathrm{LN}(\mathbf{h}^L_{\texttt{[EOS]}})$, followed by $\ell_2$ normalization.
Following Gandelsman~\etal~\cite{gandelsman2024interpreting}, who project intermediate ViT layer outputs through CLIP's final head to interpret per-layer contributions, we extend this to the text encoder by defining $\mathbf{f}^l_T = W_T\,\mathrm{LN}(\mathbf{h}^l_{\texttt{[EOS]}})$ (with $\ell_2$ normalization), so that $\mathbf{f}^L_T \equiv \mathbf{f}_T$.
Because $W_T$ was trained to maximize cosine similarity between $\mathbf{f}^L_T$ and $\mathbf{f}_I$, the score $\cos(\mathbf{f}^l_T, \mathbf{f}_I)$ at intermediate layers measures the degree to which layer~$l$ has already specialized for cross-modal alignment.

For a negated caption $T^-$ (\eg, ``not a dog'') and its affirmative counterpart $T^+$ (\eg, ``a dog''), CLIP produces $\cos(\mathbf{f}_{T^+}, \mathbf{f}_{T^-}) \approx 1$, meaning that the embeddings are nearly identical despite opposite semantics.
Kang~\etal~\cite{kang2025clip} prove that this is a geometric constraint of the shared space, not merely a training failure.
However, our analysis (\cref{sec:lcd}) reveals that compositional awareness \emph{does} exist in the frozen encoder's intermediate layers but is overwritten by the final layers.

\begin{figure}[t]
  \centering
  \includegraphics[width=0.85\linewidth]{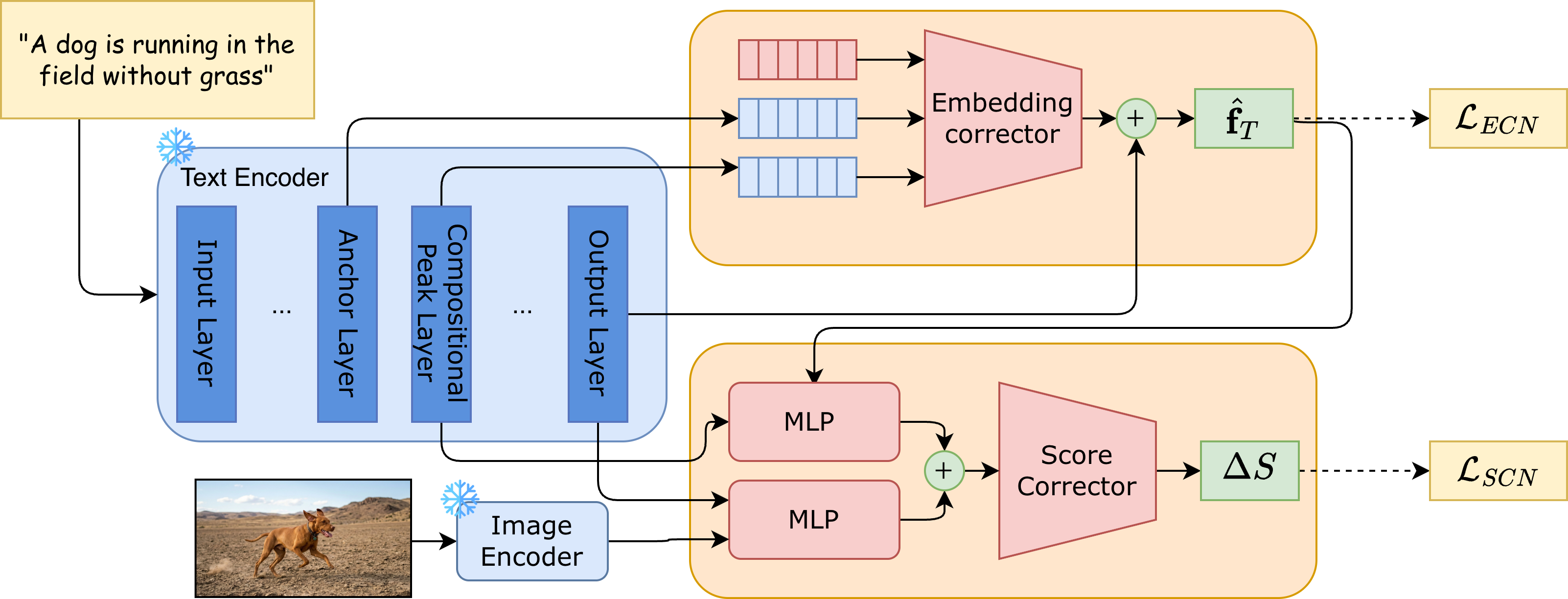}
  \caption{
\textbf{Method overview.} With all CLIP parameters frozen, we extract hidden states at three layers identified by the LCD analysis (\cref{sec:lcd}): anchor $l_a$, compositional peak $l_p$, and final layer $L$. The ECN (\cref{sec:peakpatch}) attends over the full token sequence at $l_p$ with a learned query, fuses the result with peak- and anchor-layer embeddings, and predicts a deviation vector $\boldsymbol{\delta}$ added to the collapsed layer-$L$ embedding. The SCN (\cref{sec:peakpatch}) contrasts the corrected embedding $\hat{\mathbf{f}}_T$ with the $l_p$ representation and the image embedding $\mathbf{f}_I$ to predict a bounded scalar correction $\Delta s$ to the cosine similarity. Both modules are trained jointly (\cref{sec:training}) while all CLIP parameters remain frozen.
  }
  \label{fig:method}
\end{figure}

\subsection{Layer-wise Compositional Divergence}
\label{sec:lcd}

To formalize \emph{Representational Collapse}, we define two complementary per-layer metrics:

\noindent\textbf{Compositional divergence.}
Given a diagnostic set $\mathcal{D} = \{(I_i, T^+_i, T^-_i)\}$ of image--caption triples, where $T^+$ is the affirmative caption and $T^-$ its negated counterpart, we measure how well each layer separates their projected \texttt{[EOS]} representations:
\begin{equation}
  D_{\mathrm{LCD}}(l) = \frac{1}{|\mathcal{D}|} \sum_{i} \Bigl(1 - \cos\bigl(\mathbf{f}^l_T(T^+_i),\; \mathbf{f}^l_T(T^-_i)\bigr)\Bigr).
  \label{eq:lcd}
\end{equation}

\noindent\textbf{Visual alignment.}
We complement $D_{\mathrm{LCD}}$ with a visual alignment score that measures how closely each layer's projected \texttt{[EOS]} representation matches the paired image embedding:
\begin{equation}
  A(l) = \frac{1}{|\mathcal{D}|} \sum_{i} \cos\bigl(\mathbf{f}^l_T(T^+_i),\, \mathbf{f}_{I_i}\bigr).
  \label{eq:visual_align}
\end{equation}
Because both $\mathbf{f}^l_T$ and $\mathbf{f}_I$ lie in the shared embedding space, the cosine similarity is well-defined at every layer; $A(l)$ is not a claim about inherent visual content at layer~$l$, but a measure of how strongly the contrastive alignment pressure has reshaped that layer's output.

\noindent\textbf{The collapse trajectory.}
\Cref{fig:motivation} plots both metrics across the $L$ transformer layers, revealing three phases.
In the early and middle layers, $D_{\mathrm{LCD}}$ rises steadily as the encoder builds syntactic structure, while $A(l)$ remains low.
In the upper-middle layers, $A(l)$ increases sharply as the encoder begins to specialize for cross-modal matching.
Once alignment dominates, $D_{\mathrm{LCD}}$ drops and the final layers collapse compositional distinctions into a syntax-blind, visually aligned representation.

This collapse is a direct consequence of the InfoNCE~\cite{oord2018representation} training objective.
Because negation is largely absent from web-crawled image--caption data, the contrastive loss is dominated by object-level semantics: the final layers learn to maximize visual alignment based on \emph{what} objects are present, with no gradient signal to preserve the distinction between ``a dog'' and ``not a dog.''
The compositional structure built by the middle layers is therefore overwritten as the encoder specializes for object-driven matching.
We verify that this pattern is negation-specific through a control experiment in \cref{sec:analysis}.

This trajectory identifies three functionally distinct zones that motivate our layer extraction strategy (\cref{fig:method}):

\noindent \textbf{Anchor zone (layer~$l_a$).} $D_{\mathrm{LCD}}$ is rising but $A(l)$ has not yet begun its steep ascent, providing a stable linguistic baseline uncontaminated by the visual prior.

\noindent \textbf{Compositional peak (layer~$l_p$).} $D_{\mathrm{LCD}}$ reaches its maximum, the point at which syntactic and visual signals coexist before coming into conflict.

\noindent\textbf{Collapse zone (layers~$l_p{+}1$ through~$L$).} $A(l)$ dominates and $D_{\mathrm{LCD}}$ drops sharply; by layer~$L$, the \texttt{[EOS]} embedding is visually aligned but syntax-blind.

\noindent\textbf{Deterministic layer selection.}
Both layers are read directly off the LCD curve rather than tuned: the compositional peak is $l_p = \arg\max_{\ell} D_{\mathrm{LCD}}(\ell)$, and the anchor $l_a$ is the latest pre-peak layer still in the rising-$D_{\mathrm{LCD}}$, pre-alignment zone---the closest stable baseline to $l_p$, which shares the most processing context with it.
Computing $D_{\mathrm{LCD}}$ requires only forward passes over a small diagnostic set and takes minutes, so layer selection is a one-time characterization, not a hyperparameter search.
The same rule transfers across backbones: re-running the analysis on each encoder relocates $l_p$ automatically, with no manual intervention (\cref{sec:cross_pretrain} reports a peak shift to Layer~4 for a LAION-2B-pretrained encoder, with comparable gains).

\subsection{\name}
\label{sec:peakpatch}
The compositional peak retains negation information but lacks visual alignment, while the final layer is visually aligned but syntax-blind; an effective correction must bridge both representations.
\name{} does so through two lightweight modules that operate on frozen CLIP features (\cref{fig:method}).
The Embedding Correction Network (ECN) extracts a negation-specific deviation from the peak layer and injects it into the final text embedding, producing a corrected representation suitable for retrieval and generation.
However, embedding correction alone may not suffice for discriminative tasks such as MCQ, where the decision hinges on the \emph{relative} ranking of $K$ candidate scores against a single image rather than on absolute embedding quality.
A complementary Score Correction Network (SCN) therefore predicts a bounded scalar offset to the cosine similarity, allowing fine-grained adjustment of pairwise comparisons that the embedding shift cannot fully resolve.

\noindent\textbf{Embedding Correction Network (ECN).}
\label{sec:ecn}
The ECN corrects the collapsed layer-$L$ embedding by recovering the negation signal from the compositional peak (\cref{fig:method}, top branch).
It predicts a deviation vector $\boldsymbol{\delta} \in \mathbb{R}^d$ that shifts the final embedding to separate negated from affirmative descriptions:
\begin{equation}
  \hat{\mathbf{f}}_T = \frac{\mathbf{f}_T + \alpha \cdot \boldsymbol{\delta}}{\|\mathbf{f}_T + \alpha \cdot \boldsymbol{\delta}\|_2},
  \label{eq:ecn_correct}
\end{equation}
where $\alpha$ is a learned scalar controlling the correction magnitude.

The deviation $\boldsymbol{\delta}$ is computed in two stages.
All intermediate hidden states are first passed through CLIP's final layer normalization ($\mathrm{LN}_{\mathrm{final}}$) but \emph{not} the text projection $W_T$, preserving the full token-level structure while ensuring numerical compatibility across layers; for notational brevity, we continue to write $\mathbf{H}^l$ and $\mathbf{h}^l$ for these layer-normalized representations in the equations below.
In the first stage, a learned query $\mathbf{q} \in \mathbb{R}^d$ attends over the full token sequence at the peak layer via multi-head cross-attention~\cite{vaswani2017attention} to produce a negation-aware summary:
\begin{equation}
  \mathbf{a} = \mathrm{LN}\!\bigl(\mathrm{CrossAttn}(\mathbf{q},\, \mathbf{H}^{l_p},\, \mathbf{H}^{l_p})\bigr) \in \mathbb{R}^d,
  \label{eq:crossattn}
\end{equation}
where $\mathrm{LN}$ denotes a separate learned layer normalization.
Because the query is learned end-to-end, it discovers negation-relevant token positions automatically, handling negation through a single mechanism without requiring a syntactic parser.

In the second stage, the cross-attention summary is concatenated with the \texttt{[EOS]} tokens from the peak and anchor layers and a mean-pooled representation of the peak layer, then passed through a bottleneck MLP:
\begin{equation}
  \boldsymbol{\delta} = \mathrm{MLP}_{\mathrm{ECN}}\!\bigl([\mathbf{h}^{l_p}_{\texttt{[EOS]}};\, \mathbf{a};\, \mathbf{h}^{l_a}_{\texttt{[EOS]}};\, \bar{\mathbf{h}}^{l_p}]\bigr) \in \mathbb{R}^d,
  \label{eq:deviation}
\end{equation}
where $\bar{\mathbf{h}}^{l_p}$ is the mean-pooled token representation at the peak layer and $\mathrm{MLP}_{\mathrm{ECN}}: \mathbb{R}^{4d} \to \mathbb{R}^d$.
The anchor token $\mathbf{h}^{l_a}_{\texttt{[EOS]}}$ serves as a stable reference from before alignment pressure distorts the representation, while $\bar{\mathbf{h}}^{l_p}$ provides a global summary of the peak layer's full token sequence, complementing the query-focused cross-attention output.
Since $\hat{\mathbf{f}}_T$ is $\ell_2$-normalized (\cref{eq:ecn_correct}), it preserves CLIP's cosine similarity interface and can serve as a drop-in replacement for retrieval and generation tasks.

\noindent\textbf{Score Correction Network (SCN).}
\label{sec:scn}
While the ECN operates in embedding space, the SCN provides a complementary correction at the score level (\cref{fig:method}, bottom branch).
For discriminative tasks such as multiple-choice question answering, the SCN predicts a bounded scalar adjustment to the cosine similarity between an image and a candidate caption.

The SCN reads from the compositional peak region, where negation separability is high but the representation is not yet dominated by alignment pressure.
This provides a compositional cue complementary to the token-level features the ECN extracts, making the score correction more robust on ambiguous inputs.

The SCN aggregates three sources of information.
A \emph{text context} encoder maps the concatenation of the projected peak-layer embedding $\mathbf{f}^{l_p}_T$ and the ECN-corrected embedding $\hat{\mathbf{f}}_T$ to a compact representation $\mathbf{t} \in \mathbb{R}^{d_s}$ (where $d_s \ll d$ is the SCN's internal dimension), capturing the discrepancy between the compositionally aware peak representation and the corrected output.
A \emph{cross-modal context} encoder maps the element-wise product $\mathbf{f}_I \odot \hat{\mathbf{f}}_T$ to $\mathbf{c} \in \mathbb{R}^{d_s}$, encoding how the corrected text relates to the image.
Finally, two scalar \emph{similarity features} $\mathbf{s} = [\cos(\mathbf{f}^{l_p}_T,\, \mathbf{f}_I);\; \cos(\hat{\mathbf{f}}_T,\, \mathbf{f}_I)]$ directly compare the peak-layer and corrected similarities; disagreement between these two scores signals that negation information was lost during the collapse.
These three streams are concatenated and mapped to a bounded scalar correction:
\begin{equation}
  \Delta s = \alpha_{\max} \cdot \tanh\!\bigl(\mathrm{MLP}_{\mathrm{SCN}}([\mathbf{s};\, \mathbf{t};\, \mathbf{c}])\bigr),
  \label{eq:scn}
\end{equation}
where the $\tanh$ bounds the output to $[-\alpha_{\max},\, \alpha_{\max}]$, preventing the SCN from overriding the embedding-level signal.
The final corrected similarity is:
\begin{equation}
  \hat{s}(I, T) = \cos\!\bigl(\mathbf{f}_I,\, \hat{\mathbf{f}}_T\bigr) + \Delta s.
  \label{eq:combined}
\end{equation}
Because the SCN takes $\hat{\mathbf{f}}_T$ rather than the original layer-$L$ embedding as input, embedding-level corrections propagate into the score-level module, coupling the two streams end-to-end.

\noindent\textbf{Inference.}
At test time, \name{} applies the ECN and SCN uniformly to every input: it performs no polarity classification, gating, or parsing to decide whether a caption is negated, so affirmative and negated queries pass through the identical correction path.
The ECN always returns a drop-in $\ell_2$-normalized embedding $\hat{\mathbf{f}}_T$ and the SCN always returns a bounded offset $\Delta s$; on affirmative inputs the learned corrections are small, preserving standard retrieval (\cref{sec:analysis}) and zero-shot classification (\cref{sec:zeroshot}).

\subsection{Joint Training}
\label{sec:training}
Both modules are trained jointly end-to-end while all CLIP parameters remain frozen.

\noindent\textbf{ECN objective.}
The ECN is trained with a symmetric (image-to-text + text-to-image) InfoNCE loss~\cite{oord2018representation}.
For a batch of $B$ images, each paired with an affirmative and a negated caption, the contrastive denominator sums over all $2B$ corrected embeddings, so negated captions act as hard negatives that the loss explicitly pushes apart.
We denote this loss $\mathcal{L}_{\mathrm{ECN}}$.

\noindent\textbf{SCN objective.}
The SCN is trained with a $K$-way softmax cross-entropy loss over multiple-choice questions, each consisting of one correct caption and $K{-}1$ negated distractors.
An $\ell_2$ penalty on the correction magnitude $(\Delta s)^2$, averaged over $K$ options within each MCQ, regularizes the SCN to keep adjustments small.
We denote this loss $\mathcal{L}_{\mathrm{SCN}}$.

\noindent\textbf{Joint optimization.}
The total objective combines both losses:
\begin{equation}
  \mathcal{L} = \mathcal{L}_{\mathrm{ECN}} + \lambda_{\mathrm{SCN}} \,\mathcal{L}_{\mathrm{SCN}}.
  \label{eq:loss_total}
\end{equation}
Because the SCN operates on ECN-corrected embeddings, gradients from $\mathcal{L}_{\mathrm{SCN}}$ flow back into the ECN, creating a coupled training dynamic: the ECN must produce embeddings that are useful for both retrieval (via $\mathcal{L}_{\mathrm{ECN}}$) and discrimination (via $\mathcal{L}_{\mathrm{SCN}}$).
We use separate learning rates for the two modules to balance the contrastive and discriminative objectives.
\section{Experiments}
\label{sec:experiments}

\subsection{Setup}
\label{sec:setup}


\noindent\textbf{Architecture and training.}
We build on CLIP ViT-B/32~\cite{radford2021clip} ($d{=}512$).
The ECN and SCN add ${\sim}$4.7M and ${\sim}$0.5M parameters respectively, totaling 5.2M trainable parameters (3.5\% of the frozen backbone).
Training data is constructed from CC12M~\cite{changpinyo2021cc12m} following the negation generation protocol of NegBench~\cite{alhamoud2025negbench}: for each original caption, LLaMA~3.1-8B~\cite{grattafiori2024llama} generates a semantically negated counterpart by inserting explicit negation (e.g., ``no,'' ``not,'' ``without''), producing ${\sim}$1.06M image--caption pairs for the ECN contrastive objective and ${\sim}$313K four-option MCQ samples for the SCN.
Both modules are trained jointly for 10 epochs with AdamW~\cite{loshchilov2019adamw} (weight decay $10^{-2}$, gradient clipping at norm~1.0), cosine annealing, and separate learning rates ($10^{-5}$ for the ECN, $10^{-4}$ for the SCN).
Batch size is 1024; training takes ${\sim}$2.7~hours on a single NVIDIA A100.

\noindent\textbf{Evaluation.}
We evaluate on NegBench~\cite{alhamoud2025negbench} under two protocols:
\emph{Negation MCQ}---four-choice accuracy on COCO~\cite{lin2014coco} and VOC~2007 splits (affirmation, negation, hybrid templates); and
\emph{Negation retrieval}---text-to-image Recall@1/5 with negated queries on COCO and MSR-VTT~\cite{xu2016msrvtt}.
We also evaluate on the \emph{text-to-image generation} benchmark of Park~\etal~\cite{garg2025negationclip}: ECN-corrected embeddings are fed into a frozen GALIP~\cite{tao2023galip} generator on 107 negation prompts; Gemma-3-27B~\cite{gemma2025} judges whether the affirmative object is present (Aff) and the negated attribute absent (Neg); combined (Comb\,=\,Aff\,$\times$\,Neg) counts a sample as correct only when both hold.
Zero-shot classification on CIFAR-100 is reported in \cref{sec:zeroshot}.

\noindent\textbf{Baselines.}
\emph{Encoder fine-tuning}: NegCLIP~\cite{yuksekgonul2023negclip}, CoN-CLIP~\cite{singh2025conclip}, NegationCLIP~\cite{garg2025negationclip}, and CLIP/NegCLIP\,+\,NF~\cite{alhamoud2025negbench}.
\emph{Post-hoc}: DCSM~\cite{kang2025clip} (dense token-to-patch matching, ViT-B/16).

\subsection{Main Results}
\label{sec:results}

\noindent\textbf{Negation MCQ.}
\Cref{tab:mcq} reports multiple-choice accuracy on the COCO and VOC splits of NegBench.
Baseline CLIP scores only 39.2\% on COCO, with 70.0\% on affirmative captions but 6.6\% on negated ones.
The best fine-tuning method (NegCLIP\,+\,NF) reaches 56.5\%; among post-hoc methods, DCSM achieves 48.6\%.
\name{} achieves \textbf{74.3\%} on COCO (+35.1 over CLIP, +17.8 over the best fine-tuning baseline) with a large negation gain (6.6\%$\to$63.2\%).
On VOC, \name{} reaches 65.5\%, surpassing all fine-tuning baselines.

\begin{table}[t]
  \caption{\textbf{Negation MCQ accuracy (\%) on NegBench}~\cite{alhamoud2025negbench}.
    Aff, Neg, and Hyb denote affirmation, negation, and hybrid template types;
    Avg is computed over all samples across template types.
    All methods use CLIP ViT-B/32~\cite{radford2021clip} unless noted. 
  }
  \label{tab:mcq}
  \centering
  \scriptsize
  \begin{tabular*}{\linewidth}{@{\extracolsep{\fill}}l cccc cccc}
    \toprule
    & \multicolumn{4}{c}{COCO} & \multicolumn{4}{c}{VOC} \\
    \cmidrule(lr){2-5} \cmidrule(l){6-9}
    Method & Aff & Neg & Hyb & Avg & Aff & Neg & Hyb & Avg \\
    \midrule
    CLIP~\cite{radford2021clip}
      & 70.0 & 6.6 & 38.4 & 39.2
      & 80.9 & 3.0 & 58.0 & 37.9 \\
    \midrule
    \multicolumn{9}{@{}l}{\textit{Encoder fine-tuning}} \\[2pt]
    NegCLIP~\cite{yuksekgonul2023negclip} {\scriptsize[ICLR'23]}
      & 49.2 & 13.9 & 16.3 & 26.8
      & 70.5 & 4.6 & 42.3 & 30.2 \\
    CoN-CLIP~\cite{singh2025conclip} {\scriptsize[WACV'25]}
      & 15.6 & 32.9 & 25.3 & 24.4
      & 24.8 & 23.2 & 56.7 & 38.2 \\
    CLIP\,+\,NF~\cite{alhamoud2025negbench} {\scriptsize[CVPR'25]}
      & 73.1 & 33.2 & 54.7 & 54.2
      & 85.0 & 31.7 & 79.5 & 60.1 \\
    NegCLIP\,+\,NF~\cite{alhamoud2025negbench} {\scriptsize[CVPR'25]}
      & 81.0 & 25.9 & 60.1 & 56.5
      & 81.0 & 21.1 & \textbf{83.7} & 58.2 \\
    \midrule
    \multicolumn{9}{@{}l}{\textit{Post-hoc correction }} \\[2pt]
    DCSM$^\dagger$~\cite{kang2025clip} {\scriptsize[ICCV'25]}
      & 71.2 & 6.6 & 68.0 & 48.6
      & 68.5 & 5.4 & 73.1 & 49.0 \\
    \name{}
      & \textbf{98.1} & \textbf{63.2} & \textbf{60.7} & \textbf{74.3}
      & \textbf{99.7} & \textbf{57.9} & 62.2 & \textbf{65.5} \\
    \bottomrule
  \end{tabular*}

  \vspace{2pt}
  {\raggedright\scriptsize
    $^\dagger$\,Uses ViT-B/16 backbone.
    NF\,=\,NegFull~\cite{alhamoud2025negbench} fine-tuning data.}
\end{table}

\noindent\textbf{Negation retrieval.}
\Cref{tab:retrieval} reports text-to-image retrieval with negated queries.
In the fully OOD setting (trained without COCO), \name{} outperforms all fine-tuning baselines across all metrics despite training only 5.2M parameters \vs ${\sim}$151M for full encoder fine-tuning.
In-domain, \name{} reaches 37.1\% R@1 and 64.3\% R@5 on COCO, closing much of the gap to NegCLIP\,+\,NF (41.3\% R@1) while modifying only 3.5\% of the backbone.
The remaining gap is expected: encoder fine-tuning reshapes the joint space for both modalities, whereas our correction operates only on text embeddings.

\begin{table}[t]
  \caption{\textbf{Negation retrieval (Text$\to$Image, \%) on NegBench}~\cite{alhamoud2025negbench}.
    R@$k$: Recall@$k$ on negated queries.
    All methods use CLIP ViT-B/32~\cite{radford2021clip} unless noted.
    FT\,=\,encoder fine-tuning; PH\,=\,post-hoc (frozen backbone).
  }
  \label{tab:retrieval}
  \centering
  \scriptsize
  \begin{tabular*}{\linewidth}{@{\extracolsep{\fill}}l c r cc cc}
    \toprule
    & & & \multicolumn{2}{c}{COCO} & \multicolumn{2}{c}{MSR-VTT} \\
    \cmidrule(lr){4-5} \cmidrule(l){6-7}
    Method & & \#P & R@1 & R@5 & R@1 & R@5 \\
    \midrule
    CLIP~\cite{radford2021clip}
      & -- & -- & 25.0 & 47.9 & 23.8 & 45.9 \\
    \midrule
    \multicolumn{7}{@{}l}{\textit{Trained on COCO (COCO\,=\,in-domain)}} \\[2pt]
    NegCLIP~\cite{yuksekgonul2023negclip} {\scriptsize[ICLR'23]}
      & FT & 151M & 41.0 & 68.6 & 28.0 & 50.2 \\
    NegationCLIP~\cite{garg2025negationclip} {\scriptsize[ICCV'25]}
      & FT & 151M & 38.6 & 65.8 & \textbf{29.3} & \textbf{53.8} \\
    NegCLIP\,+\,NF~\cite{alhamoud2025negbench} {\scriptsize[CVPR'25]}
      & FT & 151M & \textbf{41.3} & \textbf{69.0} & 29.2 & 51.5 \\
    DCSM$^\dagger$~\cite{kang2025clip} {\scriptsize[ICCV'25]}
      & PH & 3.0M & 10.6 & 28.6 & 19.4 & 41.2 \\
    \name{}
      & PH & 5.2M & 37.1 & 64.3 & 26.2 & 49.0 \\
    \midrule
    \multicolumn{7}{@{}l}{\textit{Trained w/o COCO (fully OOD)}} \\[2pt]
    CoN-CLIP~\cite{singh2025conclip} {\scriptsize[WACV'25]}
      & FT & 151M & 25.7 & 50.1 & 23.3 & 45.4 \\
    CLIP\,+\,NF~\cite{alhamoud2025negbench} {\scriptsize[CVPR'25]}
      & FT & 151M & 30.4 & 55.0 & 28.4 & 51.6 \\
    \name{}
      & PH & 5.2M & \textbf{31.4} & \textbf{56.9} & \textbf{29.1} & \textbf{53.7} \\
    \bottomrule
  \end{tabular*}

  \vspace{2pt}
  {\raggedright\scriptsize
    $^\dagger$\,Uses ViT-B/16 backbone. \quad
    NF\,=\,NegFull~\cite{alhamoud2025negbench} fine-tuning data.\par}
\end{table}

\noindent\textbf{Text-to-image generation.}
\Cref{tab:t2i} evaluates ECN-corrected embeddings in a downstream generation pipeline.
The corrected embeddings raise the negation score from 29.2\% to 47.6\% (+18.4) while maintaining affirmative accuracy (98.1\%), yielding a combined score of 45.8\% \vs 28.3\% for CLIP---surpassing NegationCLIP~\cite{garg2025negationclip} (44.5\%) without modifying encoder weights.

\begin{table}[t]
  \centering
  \begin{minipage}[t]{0.48\linewidth}
    \caption{\textbf{Text-to-image generation with negated prompts} (\%). Image generated with frozen GALIP~\cite{tao2023galip} generator on 107 prompts~\cite{garg2025negationclip}, correctness judged by Gemma-3-27B~\cite{gemma2025}.}
    \label{tab:t2i}
    \vspace{2pt}
    \centering
    \scriptsize
    \setlength{\tabcolsep}{0pt}
    \begin{tabular*}{\linewidth}{@{\extracolsep{\fill}}l ccc}
      \toprule
      Method & Aff & Neg & Comb \\
      \midrule
      CLIP~\cite{radford2021clip}
        & 97.4 & 29.2 & 28.3 \\
        \midrule
      NegCLIP~\cite{yuksekgonul2023negclip}
        & \textbf{98.8} & 24.5 & 23.7 \\
      NegationCLIP~\cite{garg2025negationclip}
        & \textbf{98.8} & 45.2 & 44.5 \\
      \name{} (ECN only)
        & 98.1 & \textbf{47.6} & \textbf{45.8} \\
      \bottomrule
    \end{tabular*}
  \end{minipage}
  \hfill
  \begin{minipage}[t]{0.48\linewidth}
    \centering
    \caption{Generalization across architectures (MCQ Avg, \%). \name{} consistently improves negation understanding across diverse VLM backbones without architecture-specific tuning.}
    \label{tab:generalization}
    \vspace{2pt}
    \scriptsize
    \setlength{\tabcolsep}{0pt}
    \begin{tabular*}{\linewidth}{@{\extracolsep{\fill}}l cc cc}
      \toprule
      & \multicolumn{2}{c}{Base} & \multicolumn{2}{c}{+ \name{}} \\
      \cmidrule(lr){2-3} \cmidrule(l){4-5}
      Architecture & COCO & VOC & COCO & VOC \\
      \midrule
      ViT-B/32~\cite{radford2021clip}
        & 39.2 & 37.9 & \textbf{74.3} & \textbf{65.5} \\
      ViT-L/14~\cite{radford2021clip}
        & 40.6 & 38.0 & \textbf{63.8} & \textbf{52.5} \\
      SigLIP~\cite{zhai2023siglip}
        & 28.9 & 30.8 & \textbf{66.4} & \textbf{55.3} \\
      \bottomrule
    \end{tabular*}
  \end{minipage}
\end{table}
\subsection{Ablation Study}
\label{sec:ablation}

\begin{figure}[t]
  \centering
  \begin{minipage}[t]{0.48\linewidth}
    \vspace{0pt}
    \centering
    \scriptsize
    \setlength{\tabcolsep}{0pt}
    \begin{tabular*}{\linewidth}{@{\extracolsep{\fill}}l cc c}
      \toprule
      & \multicolumn{2}{c}{MCQ} & Retrieval \\
      \cmidrule(lr){2-3} \cmidrule(l){4-4}
      Variant & Neg & Avg & R@5 \\
      \midrule
      CLIP (baseline)
        & 6.6 & 39.2 & 47.9 \\
      \midrule
      ECN only
        & 16.5 & 51.2 & 58.2 \\
      SCN only
        & \textbf{67.3} & 71.6 & 48.1 \\
     w/o anchor ($l_a$)
        & 61.7 & 68.7 & 60.5 \\
      Detached SCN
        & 57.3 & 67.1 & \textbf{64.8} \\
        \midrule
      \name{} (joint)
        & 63.2 & \textbf{74.3} & 64.3 \\

      \bottomrule
    \end{tabular*}
    \captionof{table}{\textbf{Ablation study} on COCO (\%).
      Top: component contribution. Bottom: design choices.
      MCQ: negation accuracy and overall average.
      Ret: negation retrieval R@5.}
    \label{tab:ablation}
  \end{minipage}
  \hfill
  \begin{minipage}[t]{0.48\linewidth}
    \vspace{0pt}
    \centering
    \includegraphics[width=\linewidth]{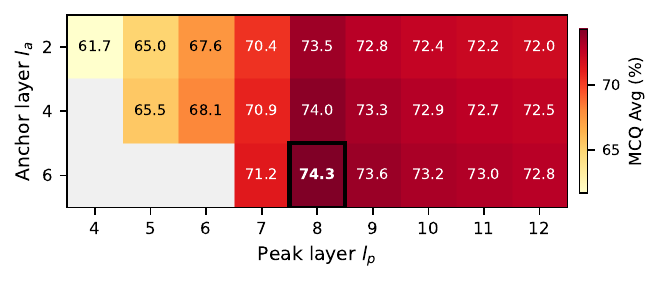}
    \caption{\textbf{ECN layer selection.} COCO MCQ Avg as a function of peak layer $l_p$ and anchor layer $l_a$. Accuracy peaks at $l_p{=}8$, coinciding with the LCD maximum (\cref{sec:lcd}), and is robust to the anchor choice.}
    \label{fig:layer_choice}
  \end{minipage}
 
\end{figure}

\noindent\textbf{Component contribution.}
\Cref{tab:ablation} isolates the contribution of each module on COCO.
The standalone ECN improves MCQ Avg from 39.2\% to 51.2\% and retrieval R@5 from 47.9\% to 58.2\%, showing that embedding-level correction benefits both tasks.
The standalone SCN provides a larger MCQ gain (71.6\%) but barely improves retrieval (48.1\%), as score-level correction does not alter the embedding space.
Joint training yields the best overall system (74.3\% MCQ, 64.3\% R@5), with the ECN and SCN co-adapting through shared gradients.

\noindent\textbf{Design choices.}
\Cref{tab:ablation} also ablates two key architectural decisions.
Removing the anchor layer ($l_a$) reduces accuracy by $-$5.6 (74.3\%$\to$68.7\%), confirming that the stable baseline from the anchor zone is critical for measuring compositional change.
Detaching the SCN from ECN gradients reduces accuracy by $-$7.2 (74.3\%$\to$67.1\%), showing that the cooperative training dynamic between the two modules is essential.

\noindent\textbf{Layer selection.}
\Cref{fig:layer_choice} sweeps the ECN peak layer $l_p$ and anchor layer $l_a$ across CLIP's text-encoder layers.
Accuracy peaks at $l_p{=}8$, the same layer where the LCD trajectory reaches its maximum (\cref{sec:lcd}), and decreases for both earlier and later choices.
Performance is robust to the anchor choice, varying by $<$1\,pp across $l_a \in \{2,4,6\}$ for a fixed peak, confirming that the peak layer is the critical design decision and that the LCD analysis provides a principled criterion for selecting it.


\subsection{Analysis}
\label{sec:analysis}

\noindent\textbf{Generalization across architectures.}
\Cref{tab:generalization} evaluates \name{} on additional backbones.
Negation blindness persists regardless of scale or objective---CLIP ViT-L/14 (40.6\%) and SigLIP ViT-B/16 (28.9\%) match ViT-B/32 (39.2\%)---and \name{} consistently improves COCO MCQ: +35.1 on ViT-B/32, +23.2 on ViT-L/14 (to 63.8\%), and +37.5 on SigLIP (to 66.4\%).
The ViT-L/14 LCD trajectory shows an identical collapse pattern, and the layer-selection rule transfers across pretraining data and depth (OpenCLIP ViT-B/32 on LAION-2B, ViT-g/14); see \cref{sec:lcd_validation,sec:cross_pretrain}.

\noindent\textbf{Standard (non-negation) retrieval.}
Because \name{} corrects every query without gating (\cref{sec:peakpatch}), we check that it does not harm ordinary retrieval.
On standard (affirmative) MSCOCO~5K and Flickr30K~1K text-to-image retrieval (\cref{tab:standard_retrieval}), ECN-corrected embeddings slightly \emph{improve} over frozen CLIP (R@1 $+2.2$/$+2.5$), and zero-shot classification is likewise preserved (\cref{sec:zeroshot}).

\begin{table}[t]
  \caption{\textbf{Standard (non-negation) text-to-image retrieval} (\%). ECN-corrected embeddings preserve---and slightly improve---retrieval on affirmative captions.}
  \label{tab:standard_retrieval}
  \centering
  \footnotesize
  \setlength{\tabcolsep}{6pt}
  \begin{tabular}{@{}l cc cc@{}}
    \toprule
    & \multicolumn{2}{c}{MSCOCO 5K} & \multicolumn{2}{c}{Flickr30K 1K} \\
    \cmidrule(lr){2-3} \cmidrule(l){4-5}
    Method & R@1 & R@5 & R@1 & R@5 \\
    \midrule
    CLIP~\cite{radford2021clip} & 29.9 & 54.1 & 57.9 & 83.0 \\
    \name{} (ECN) & \textbf{32.1} & \textbf{57.0} & \textbf{60.4} & \textbf{83.9} \\
    \bottomrule
  \end{tabular}
\end{table}

\begin{figure}[t]
  \centering
  \includegraphics[width=0.75\linewidth]{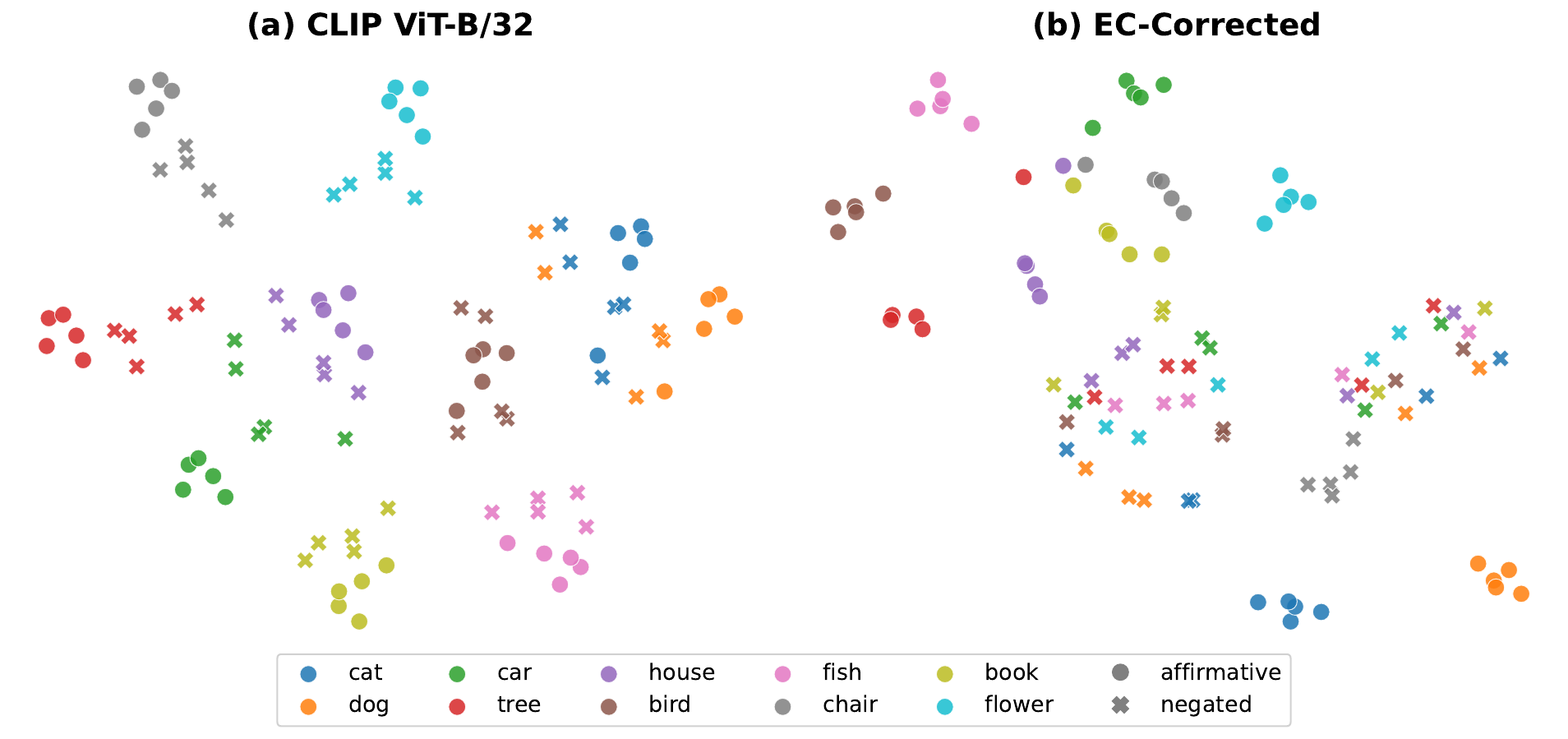}
  \caption{\textbf{t-SNE of text embeddings for 10 object categories.} Circles: affirmative captions; crosses: negated captions; colors: object categories. \emph{Left}: original CLIP embeddings (affirmative and negated overlap within each object). \emph{Right}: after ECN correction (affirmative captions stay in tight per-category clusters; negated captions separate into a looser shared region).}
  \label{fig:tsne}
\end{figure}

\noindent\textbf{Embedding space visualization.}
\Cref{fig:tsne} shows t-SNE projections of text embeddings for 10 object categories, each with 5 affirmative and 5 negated caption templates.
In the original CLIP space (\emph{left}), affirmative and negated embeddings for the same object cluster together, confirming that the encoder collapses the syntactic distinction.
After ECN correction (\emph{right}), the affirmative captions keep their tight per-category clusters while the negated captions separate from them into a distinct, looser region. We discuss negation's set-valued nature in \cref{sec:setvalued}.

\begin{figure}[t]
  \centering
  \begin{minipage}[t]{0.48\linewidth}
    \centering
    \includegraphics[width=0.9\linewidth]{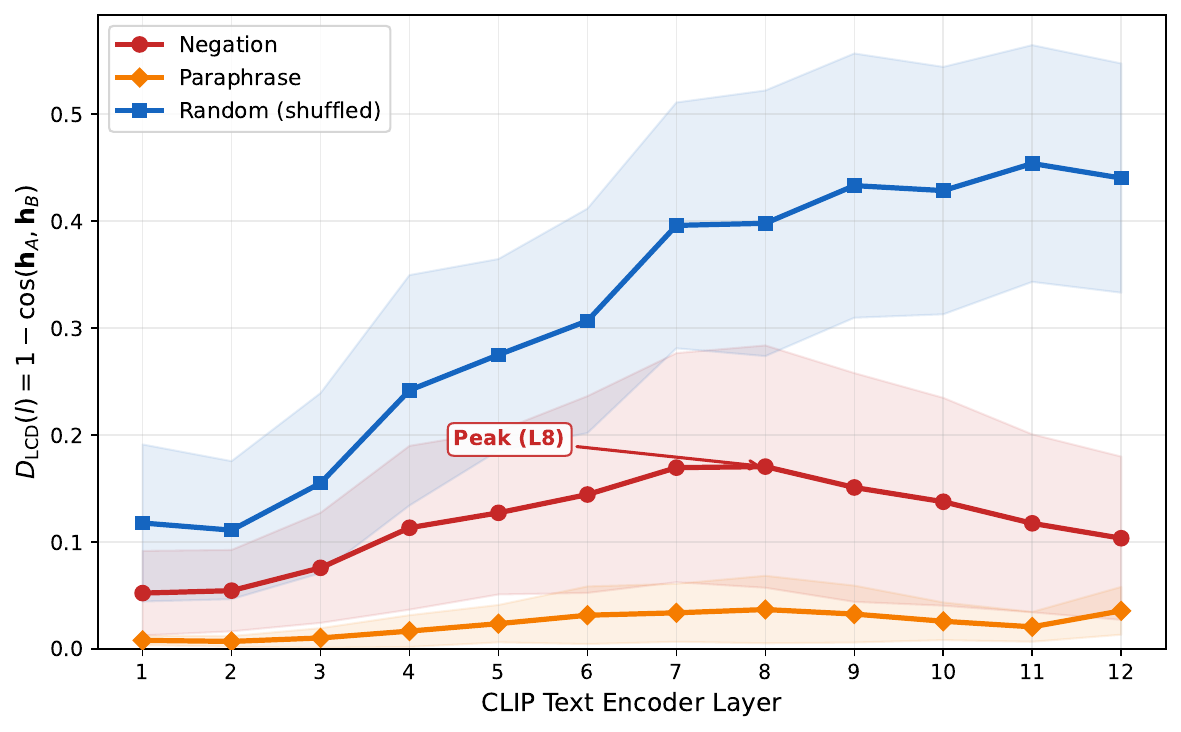}
    \caption{\textbf{LCD control experiment.} LCD is measured on 1K pairs under three conditions: negation, paraphrase, and random. Only negation exhibits the rise-and-fall pattern, confirming the signal is negation-specific.}
    \label{fig:lcd_control}
  \end{minipage}
  \hfill
  \begin{minipage}[t]{0.48\linewidth}
    \centering
    \includegraphics[width=0.9\linewidth]{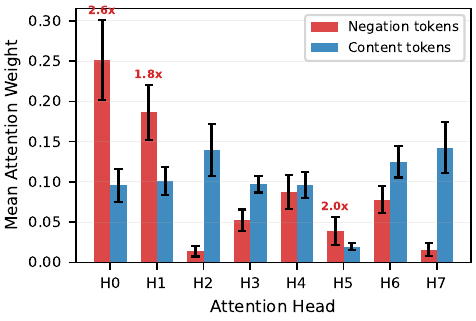}
    \caption{\textbf{ECN cross-attention by token type.} Mean attention on negation vs.\ content tokens. Heads H0, H1, and H5 attend preferentially to negation tokens (>$1.5
    \times$), compared to 8\% of CLIP heads~\cite{quantmeyer2024clip}.}
    \label{fig:ecn_attention}
  \end{minipage}
\end{figure}

\noindent\textbf{LCD control experiment.}
To verify that the LCD trajectory reflects a negation-specific phenomenon, we compare three conditions on 1K pairs (\cref{fig:lcd_control}): negation, paraphrase (synonym rewording), and random (unrelated captions).
The paraphrase curve stays near zero (meaning changes do not alter the \texttt{[EOS]} trajectory), the random curve rises monotonically, and only the negation curve exhibits the characteristic rise-and-fall, isolating Representational Collapse as unique to negation.

\noindent\textbf{Attention analysis.}
Quantmeyer~\etal~\cite{quantmeyer2024clip} found that only 8\% of CLIP's attention heads are negator-selective.
\Cref{fig:ecn_attention} measures per-head attention on negation \vs content tokens across 12 negated prompts.
Three of eight ECN heads (H0, H1, H5) allocate 1.8--2.6$\times$ more attention to negation tokens (37.5\% negation-selective); the remaining heads focus on content words, providing semantic context for \emph{what} is being negated.

\section{Conclusion}
\label{sec:conclusion}

We identified \emph{Representational Collapse} in CLIP's text encoder: intermediate layers build compositional structure separating negated from affirmative descriptions, but later layers overwrite it as they specialize for visual alignment.
Guided by this finding, we proposed \name{}, a post-hoc system that extracts the negation signal at the compositional peak through an Embedding Correction Network (ECN) and applies bounded score offsets through a Score Correction Network (SCN), trained jointly with all CLIP parameters frozen---adding 5.2M parameters (3.5\% of the backbone).
On NegBench, \name{} reaches 74.3\% COCO MCQ (+35.1 over CLIP, +17.8 over the best fine-tuning method), outperforms all fine-tuning baselines on fully out-of-distribution retrieval, transfers to text-to-image generation (+18.4 negation score), and generalizes across ViT-B/32, ViT-L/14, and SigLIP.
We discuss limitations in \cref{sec:limitations_main}.

\section{Acknowledgements}
This material is based in part upon work supported by the National Science Foundation under Grant Numbers CNS-2333487 (NSF Frontier) and CNS-2146449 (NSF CAREER). Any opinions, findings, conclusions, or recommendations expressed in this material are those of the authors and do not necessarily reflect the views of the sponsors. We thank the reviewers and area chairs for their constructive feedback, which helped improve this work.

\bibliographystyle{splncs04}
\bibliography{main}

\clearpage
\appendix
\newpage
\begin{center}
  \Large\textbf{Appendix}\\[2pt]
  \large What CLIP Knows but Cannot Say: Recovering Negation from Frozen Intermediate Features
\end{center}

\section{Limitations}
\label{sec:limitations_main}
\name{} corrects only the text side, leaving CLIP's image encoder frozen.
This is a deliberate choice---it keeps \name{} a low-risk, drop-in correction and preserves the visual representation---but it also explains the one setting where \name{} trails encoder fine-tuning: in-domain retrieval (\cref{tab:retrieval}), where methods that fine-tune both encoders co-adapt the image and text spaces to the target distribution, whereas \name{} adjusts only text.
\name{} also requires white-box access to CLIP's intermediate activations, so it does not apply to black-box deployments that expose only the final embedding or similarity score.
Finally, because negation is set-valued, \name{} recovers the affirmative/negated distinction but does not resolve a negated phrase to a specific complement concept.
Extending the correction to image embeddings and to larger backbones and broader compositional benchmarks remains future work.

\section{Set-valued Nature of Negation}
\label{sec:setvalued}
In \cref{fig:tsne},
because ``not a car'' and ``not a tree'' denote highly overlapping sets of referents, a faithful representation should place different negated phrases closer to one another than their affirmative counterparts; the negated side is therefore \emph{expected} to show weaker per-category structure than the affirmative side, without collapsing to a single object-specific point.
NegBench's reference embedding~\cite{alhamoud2025negbench} exhibits exactly this signature: tight per-object affirmative clusters and a more diffuse negated region.
\Cref{fig:tsne} (right) reproduces it---affirmative captions stay in tight per-category clusters while negated captions separate into a looser shared region with no point collapse---indicating that \name{} recovers the affirmative/negated distinction in line with negation semantics rather than amplifying negation at the expense of meaning.

\section{Zero-Shot Classification}
\label{sec:zeroshot}

To verify that \name{}'s corrections do not degrade general-purpose performance, we evaluate zero-shot classification on CIFAR-10, CIFAR-100~\cite{krizhevsky2009cifar}, and ImageNet~\cite{deng2009imagenet}.
We report ECN-corrected accuracy alongside the frozen CLIP baseline, since the ECN produces a drop-in replacement embedding compatible with any cosine-similarity pipeline; the SCN is designed for pairwise scoring tasks and does not apply to standard zero-shot classification.

\begin{table}[h]
  \caption{\textbf{Zero-shot classification accuracy (\%).} ECN-corrected embeddings preserve CLIP's recognition performance.}
  \label{tab:zeroshot}
  \centering
  \footnotesize
  \setlength{\tabcolsep}{5pt}
  \begin{tabular}{@{}l cc cc cc@{}}
    \toprule
    & \multicolumn{2}{c}{CIFAR-10} & \multicolumn{2}{c}{CIFAR-100} & \multicolumn{2}{c}{ImageNet} \\
    \cmidrule(lr){2-3} \cmidrule(lr){4-5} \cmidrule(l){6-7}
    Method & Top-1 & Top-5 & Top-1 & Top-5 & Top-1 & Top-5 \\
    \midrule
    CLIP~\cite{radford2021clip}
      & 88.3 & 99.2 & 62.5 & 87.0 & 47.4 & 68.1 \\
    \name{} (ECN)
      & 88.5 & 99.3 & 62.9 & 86.9 & 44.6 & 65.7 \\
    \bottomrule
  \end{tabular}
\end{table}

\Cref{tab:zeroshot} shows that ECN-corrected embeddings largely preserve zero-shot performance: Top-1 changes by $<$1\,pp on CIFAR-10/100 and $-$2.8\,pp on ImageNet, a modest cost given the +35\,pp gain on negation MCQ. Encoder fine-tuning methods such as NegCLIP\,+\,NF~\cite{alhamoud2025negbench} incur larger zero-shot drops~\cite{aggarwal2026seeing}.

\section{LCD Validation}
\label{sec:lcd_validation}

We provide two additional validations of the LCD analysis in \cref{sec:lcd}.

\noindent\textbf{Linear probe.}
We train a logistic regression classifier at each of the $L$ transformer layers to distinguish negated from affirmative captions, using the \texttt{[EOS]} embeddings from 117K pre-extracted validation pairs (80/20 train/test split).
Probe accuracy rises from 87.6\% at Layer~1 to a peak of 94.0\% at Layer~5, then monotonically declines to 87.9\% at Layer~$L$, confirming that the middle layers contain linearly separable negation information that is progressively destroyed as the encoder specializes for cross-modal alignment.

\begin{figure}[h]
  \centering
  \includegraphics[width=0.5\linewidth]{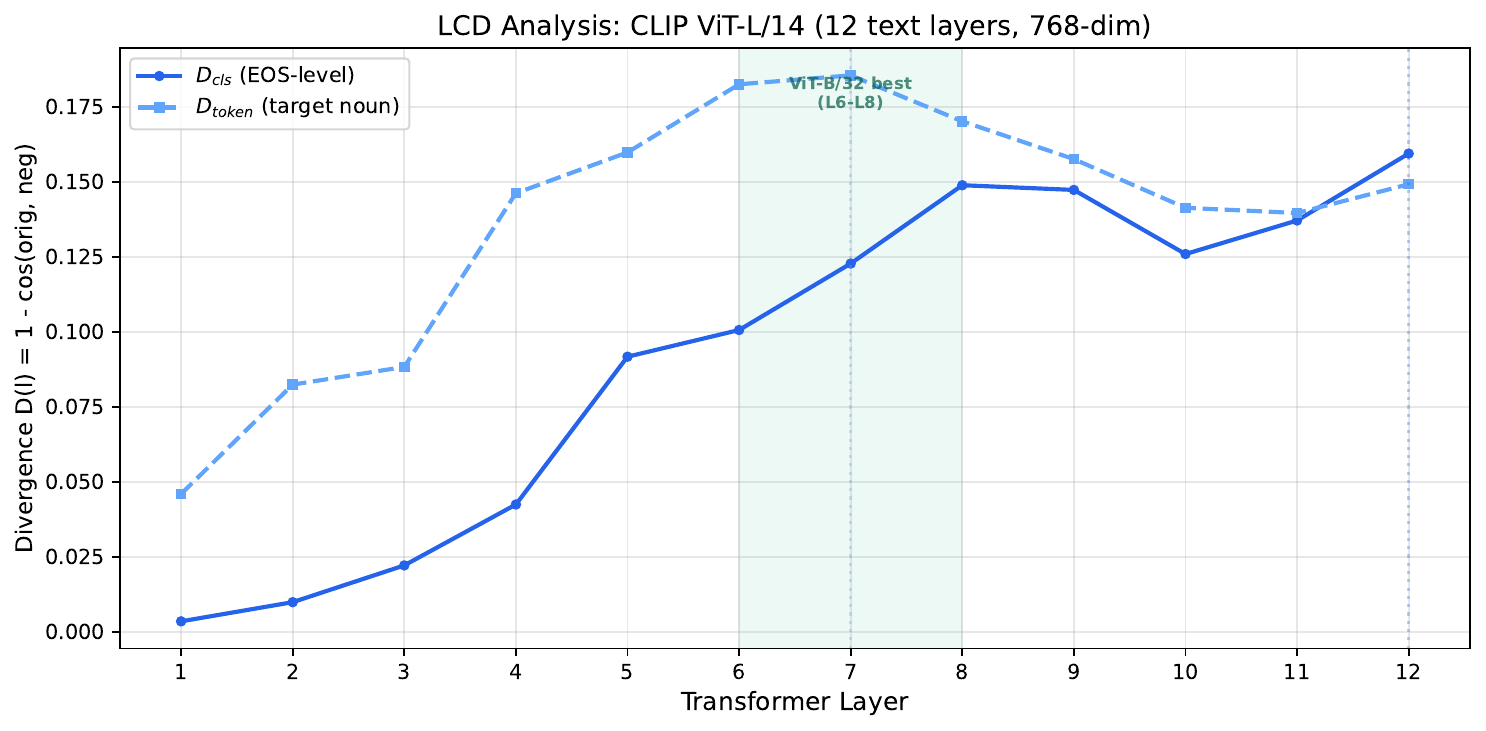}
  \caption{\textbf{LCD trajectory for CLIP ViT-L/14.} $D_{\mathrm{LCD}}$ peaks at the compositional peak $l_p$ (Layer~8), mirroring the ViT-B/32 pattern (\cref{fig:motivation}) and confirming that Representational Collapse is architecture-agnostic.}
  \label{fig:lcd_vitl14}
\end{figure}

\noindent\textbf{Generalization to ViT-L/14.}
\Cref{fig:lcd_vitl14} plots the LCD trajectory for CLIP ViT-L/14's text encoder ($L{=}12$, $d{=}768$).
The pattern is qualitatively identical to ViT-B/32: $D_{\mathrm{LCD}}$ rises through early layers, peaks at Layer~8 ($D_{\mathrm{LCD}}{=}0.149$), and then drops as the encoder specializes for visual alignment.
A token-level variant of $D_{\mathrm{LCD}}$---computed by averaging the per-position cosine distances across all token positions rather than using only the \texttt{[EOS]} token---peaks slightly earlier at Layer~7 (0.186), consistent with the ViT-B/32 pattern.
This confirms that Representational Collapse is not an artifact of the ViT-B/32 architecture but a general consequence of the contrastive training objective, motivating the same layer-extraction strategy across backbones.

\section{Cross-Pretraining and Deeper Backbones}
\label{sec:cross_pretrain}

To test whether \name{}'s layer-selection rule and corrections transfer beyond the OpenAI ViT-B/32 encoder used in our main experiments, we apply the full pipeline to encoders that differ in pretraining data and depth.
For each encoder we first re-run the LCD analysis (\cref{sec:lcd}) to locate the compositional peak $l_p$ automatically, then train \name{} with no other changes.

\noindent\textbf{Pretraining data.}
On OpenCLIP ViT-B/32 pretrained on LAION-2B~\cite{cherti2023openclip} (\vs the WIT data behind OpenAI CLIP), the LCD peak shifts from Layer~8 to Layer~4, detected by the analysis without manual intervention.
With the relocated peak, \name{} raises COCO negation accuracy from 14.0\% to 70.2\% and VOC from 5.4\% to 64.6\%---gains ($+56.2$/$+59.2$) comparable to or larger than those on OpenAI CLIP ($+56.6$/$+54.9$).

\noindent\textbf{Encoder depth.}
On OpenCLIP ViT-g/14 ($\sim$1B vision parameters, LAION-2B), the 14.2\% COCO baseline is consistent with NegBench's~\cite{alhamoud2025negbench} finding that scaling LAION-pretrained CLIP does not by itself improve negation; \name{} lifts it to 68.9\%, confirming that the correction transfers to substantially deeper encoders.

\begin{table}[h]
  \caption{\textbf{Cross-pretraining and deeper backbones.} Negation-subset MCQ accuracy (\%), CLIP $\to$ +\,\name{}. The compositional peak $l_p$ is re-detected automatically per encoder by the LCD analysis.}
  \label{tab:cross_pretrain}
  \centering
  \footnotesize
  \setlength{\tabcolsep}{6pt}
  \begin{tabular}{@{}l l cc@{}}
    \toprule
    Backbone & Pretrain & COCO Neg & VOC Neg \\
    \midrule
    ViT-B/32 & WIT      & 6.6 $\to$ \textbf{63.2} & 3.0 $\to$ \textbf{57.9} \\
    ViT-B/32 & LAION-2B & 14.0 $\to$ \textbf{70.2} & 5.4 $\to$ \textbf{64.6} \\
    ViT-g/14 & LAION-2B & 14.2 $\to$ \textbf{68.9} & 6.1 $\to$ \textbf{57.9} \\
    \bottomrule
  \end{tabular}
\end{table}

These results address three points: the selection rule transfers without re-tuning, per-backbone peak shifts are detected by LCD itself, and the framework benefits from---rather than is hindered by---per-backbone characterization.

\section{Comparison with Parser-Dependent Methods}
\label{sec:parser_comparison}

Two concurrent methods explicitly extract the negated concept from each caption at inference time, making them not directly comparable to approaches that operate on the raw caption.
We compare with both in detail.

\noindent\textbf{SpaceVLM}~\cite{ranjbar2025spacevlm} models negation through spherical cap intersections in CLIP's embedding space and requires Mistral-7B~\cite{jiang2023mistral} at inference time to parse the negation structure of each input caption.
\textbf{Aggarwal~\etal}~\cite{aggarwal2026seeing} propose a training-free approach via vector arithmetic: given a caption embedding $\mathbf{e}_c$ and the embedding of the negated concept $\mathbf{e}_{\mathrm{neg}}$, they compute $\mathbf{e}^* = \mathbf{e}_c - \lambda(\mathrm{proj}(\mathbf{e}_c, \mathbf{e}_{\mathrm{neg}}) - \mathbf{a})$, where $\mathbf{a}$ is a fixed anchor (mean embedding of semantically neutral words) and $\lambda$ a scaling factor.
This method requires a rule-based parser to identify the negated concept.
\textbf{\name{}} requires no parser at all: the ECN's learned cross-attention reads negation signals directly from CLIP's intermediate token sequences, handling all negation forms (explicit, implicit, morphological) without external parsing.

\Cref{tab:parser_mcq} compares all three methods on NegBench MCQ.

\begin{table}[t]
  \caption{\textbf{Negation MCQ accuracy (\%)} comparing parser-dependent methods with \name{} on NegBench~\cite{alhamoud2025negbench}. All methods use ViT-B/32 architecture.}
  \label{tab:parser_mcq}
  \centering
  \footnotesize
  \setlength{\tabcolsep}{3.5pt}
  \begin{tabular}{@{}l l cccc cccc@{}}
    \toprule
    & & \multicolumn{4}{c}{COCO} & \multicolumn{4}{c}{VOC} \\
    \cmidrule(lr){3-6} \cmidrule(l){7-10}
    Method & Parser & Aff & Neg & Hyb & Avg & Aff & Neg & Hyb & Avg \\
    \midrule
    CLIP~\cite{radford2021clip} & --
      & 70.0 & 6.6 & 38.4 & 39.2
      & 80.9 & 3.0 & 58.0 & 37.9 \\
    SpaceVLM$^\ddagger$~\cite{ranjbar2025spacevlm} & LLM
      & 77.4 & \textbf{71.8} & 50.0 & 66.3
      & 85.8 & \textbf{84.2} & 76.8 & \textbf{81.1} \\
    Aggarwal~\etal$^\dagger$~\cite{aggarwal2026seeing} & Rule
      & 77.8 & 64.3 & \textbf{74.8} & 72.5
      & 96.4 & 63.3 & \textbf{87.8} & 78.6 \\
    \name{} & None
      & \textbf{98.1} & 63.2 & 60.7 & \textbf{74.3}
      & \textbf{99.7} & 57.9 & 62.2 & 65.5 \\
    \bottomrule
  \end{tabular}

  \vspace{2pt}
  {\raggedright\footnotesize
    $^\ddagger$\,Requires Mistral-7B (7B params) at inference.\quad
    $^\dagger$\,Our reimplementation; uses OpenCLIP ViT-B/32 (LAION-2B).\par}
\end{table}

\noindent\textbf{MCQ results.}
On COCO MCQ, \name{} leads overall (74.3\%), outperforming SpaceVLM (66.3\%) and Aggarwal~\etal (72.5\%).
On VOC, both parser-dependent methods achieve higher accuracy (SpaceVLM 81.1\%, Aggarwal~\etal 78.6\%) than \name{} (65.5\%).
VOC contains only 20 object classes, so most negation captions follow a narrow set of patterns (\eg, ``not a cat,'' ``no bicycle'') with limited lexical variation.
In this setting, explicit parsing is highly reliable: the negated concept is almost always a single, unambiguous noun that the parser extracts correctly, yielding a precise correction.
\name{}, which learns negation signals from CLIP's intermediate representations without access to a parser, must handle the same disambiguation implicitly.
Because \name{} is trained on CC12M-derived negation pairs (\cref{sec:training_data})---a broad web-crawled distribution---it does not specialize for VOC's narrow category set.
On COCO, where 80 classes, compound objects, and more varied negation structures (\eg, multi-clause, implicit) make parsing less reliable, \name{}'s learned correction generalizes better.

\begin{table}[t]
  \centering
  \footnotesize
  \begin{minipage}[t]{0.52\linewidth}
    \centering
    \caption{\textbf{Negation retrieval (T$\to$I, \%)} including SpaceVLM.}
    \label{tab:spacevlm_retrieval}
    \setlength{\tabcolsep}{3pt}
    \begin{tabular}{@{}l cc@{}}
      \toprule
      Method & R@1 & R@5 \\
      \midrule
      CLIP~\cite{radford2021clip}
        & 25.0 & 47.9 \\
      SpaceVLM$^\ddagger$~\cite{ranjbar2025spacevlm}
        & 29.9 & 55.1 \\
      \name{}
        & \textbf{31.4} & \textbf{56.9} \\
      \bottomrule
    \end{tabular}
  \end{minipage}
  \hfill
  \begin{minipage}[t]{0.45\linewidth}
    \centering
    \caption{\textbf{T2I generation (\%)} including SpaceVLM.}
    \label{tab:spacevlm_t2i}
    \setlength{\tabcolsep}{3pt}
    \begin{tabular}{@{}l ccc@{}}
      \toprule
      Method & Aff & Neg & Comb \\
      \midrule
      CLIP~\cite{radford2021clip}
        & 97.4 & 29.2 & 28.3 \\
      SpaceVLM$^\ddagger$~\cite{ranjbar2025spacevlm}
        & \textbf{98.8} & \textbf{60.9} & \textbf{59.7} \\
      \name{} (ECN only)
        & 98.1 & 47.6 & 45.8 \\
      \bottomrule
    \end{tabular}
  \end{minipage}

  \vspace{2pt}
  {\raggedright\footnotesize
    $^\ddagger$\,Requires Mistral-7B at inference time.\par}
\end{table}

\noindent\textbf{Retrieval and generation.}
\Cref{tab:spacevlm_retrieval,tab:spacevlm_t2i} report retrieval and T2I results (Aggarwal~\etal do not evaluate T2I generation).
\name{} outperforms SpaceVLM on retrieval (R@1 31.4 \vs 29.9, R@5 56.9 \vs 55.1) despite not using an external parser.
On T2I generation, SpaceVLM leads (Neg 60.9 \vs 47.6) by leveraging Mistral-7B to parse negation structure explicitly.

\noindent\textbf{Key differences.}
\textit{Parser dependency}: SpaceVLM requires Mistral-7B at inference; Aggarwal~\etal require a rule-based parser that cannot handle implicit or morphological negation. \name{} uses no external parser.
\textit{Inference cost}: SpaceVLM adds a 7B LLM forward pass per caption; \name{} adds only 5.2M parameters (3.5\% of CLIP).
\textit{Operating space}: both baselines operate in CLIP's final embedding space, subject to the geometric impossibility of Kang~\etal~\cite{kang2025clip}; \name{} extracts features from intermediate layers and applies corrections via the SCN.
\textit{Downstream compatibility}: \name{} produces a drop-in CLIP embedding; parser-dependent methods require integrating the parser into each downstream system.

\section{External Benchmark Evaluation}
\label{sec:external_benchmarks}

To assess generalization beyond NegBench, we evaluate on two additional negation benchmarks used by NegationCLIP~\cite{garg2025negationclip}.

\noindent\textbf{Benchmarks.}
\emph{VALSE Existence}~\cite{parcalabescu2022valse} contains 505 validated pairwise samples where a caption and its foil differ by the word ``no'' (\eg, ``There are giraffes'' \vs ``There are no giraffes''); the metric is pairwise accuracy.
\emph{NegRefCOCOg}~\cite{garg2025negationclip} contains 440 region-selection samples where a negation phrase must be matched to the correct bounding-box crop from two candidates; the metric is selection accuracy.
Both benchmarks use COCO images.

NegationCLIP~\cite{garg2025negationclip} is trained on 229K pairs from COCO captions (147K) and VQAv2 (82K, also COCO images).

\begin{table}[t]
  \caption{\textbf{External benchmark evaluation (\%).} Pairwise accuracy on VALSE Existence and selection accuracy on NegRefCOCOg. \name{} results shown for ECN-only, SCN-only, and full hybrid (ECN+SCN).}
  \label{tab:external_benchmarks}
  \centering
  \footnotesize
  \setlength{\tabcolsep}{4pt}
  \begin{tabular}{@{}l l cccc c@{}}
    \toprule
    & & \multicolumn{4}{c}{\name{}} & \\
    \cmidrule(lr){3-6}
    Benchmark & CLIP & ECN & SCN & Hybrid & & NegationCLIP \\
    \midrule
    VALSE Existence & 69.5 & 79.8 & 73.3 & \textbf{80.4} & & 80.2 \\
    NegRefCOCOg     & 59.6 & 62.5 & 61.1 & \textbf{64.3} & & 64.1 \\
    \bottomrule
  \end{tabular}
\end{table}

\noindent\textbf{Results.}
\Cref{tab:external_benchmarks} reports results.
On VALSE Existence, the \name{} hybrid (80.4\%) slightly exceeds NegationCLIP (80.2\%), showing that the post-hoc correction architecture can match encoder fine-tuning even on short templated sentences.
On NegRefCOCOg, the hybrid (64.3\%) again slightly exceeds NegationCLIP (64.1\%).
Notably, the SCN contributes on this region-selection task despite not being designed for it: the hybrid outperforms ECN-only by +1.8\,pp.
Both results confirm that \name{} achieves competitive or superior performance on external benchmarks without modifying CLIP's weights.

\section{Near-Tie Stress Test}
\label{sec:near_tie}

To better understand when score-level correction helps, we construct a \emph{near-tie} subset of the NegBench COCO MCQ evaluation.
For each four-option question, we compute the frozen CLIP cosine similarities for all candidate captions and record the margin between the top-1 and top-2 scores.
We then select the hardest questions---those with the smallest margins (bottom 25\%)---where CLIP is least confident and most susceptible to negation confusion.

\begin{table}[h]
  \caption{\textbf{Near-tie stress test} on COCO MCQ (\%). The near-tie subset selects questions where frozen CLIP's top-1 and top-2 cosine scores have the smallest margin (bottom 25\%, $n{=}1{,}479$).}
  \label{tab:near_tie}
  \centering
  \footnotesize
  \setlength{\tabcolsep}{4pt}
  \begin{tabular}{@{}l cc@{}}
    \toprule
    & Full set & Near-tie \\
    Method & ($n{=}5{,}914$) & ($n{=}1{,}479$) \\
    \midrule
    CLIP~\cite{radford2021clip}  & 39.6 & 29.1 \\
    NegCLIP~\cite{yuksekgonul2023negclip} & 26.5 & 17.6 \\
    NegCLIP\,+\,NF~\cite{alhamoud2025negbench} & 56.8 & 53.5 \\
    \midrule
    ECN only         & 54.6 & 45.5 \\
    \name{} (joint)  & \textbf{67.2} & \textbf{62.3} \\
    \bottomrule
  \end{tabular}
\end{table}

\Cref{tab:near_tie} reports accuracy on the full set alongside this hard subset.
NegCLIP degrades below CLIP on this MCQ task (26.5\%), consistent with documented limitations in distinguishing semantically close alternatives~\cite{alhamoud2025negbench}.
NegCLIP\,+\,NF is a strong baseline (56.8\% full, 53.5\% near-tie) but still falls short of our joint system.
On the near-tie subset, \name{} (joint) achieves 62.3\%, dropping only 4.9\,pp from the full set, compared to a 9.1\,pp drop for the ECN alone (54.6\%$\to$45.5\%).
This confirms that the SCN's score-level correction is most valuable precisely when the embedding-level signal is insufficient to resolve close calls.

\noindent\textbf{Qualitative examples.}
\Cref{fig:near_tie_examples} shows five near-tie questions where CLIP selects the wrong answer but \name{} corrects the ranking.
In each case, the correct caption and the top-scoring distractor mention the same objects with opposite negation polarity, producing CLIP cosine scores that differ by less than 0.005.
Our joint model re-ranks the options correctly by leveraging the compositional signal from intermediate layers.

\newcommand{\neartierow}[8]{%
  \begin{minipage}[c]{0.14\linewidth}
    \centering
    \includegraphics[width=\linewidth,height=\linewidth,keepaspectratio]{#1}\\[0pt]
    {\scriptsize\textit{#2}}
  \end{minipage}\hfill
  \begin{minipage}[c]{0.84\linewidth}
    \scriptsize
    \begin{tabular}{@{}p{6.6cm} r@{\;\;}r@{}}
      & {CLIP} & {Ours} \\[-2pt]
      \colorbox{green!12}{\strut\textbf{#3}} & \texttt{#4} & \textbf{\texttt{#5}} \\[1pt]
      \colorbox{red!12}{\strut #6} & \texttt{#7} & \texttt{#8}
    \end{tabular}
  \end{minipage}%
}

\begin{figure*}[h]
  \centering

  \neartierow{}{Hybrid}%
    {(a) Contains a sports ball, but no car.}{.208}{.175}%
    {(b) Features a car, but no sports ball.}{.227}{$-$.218}

  \vspace{1pt}\hrule\vspace{1pt}

  \neartierow{}{Positive}%
    {(a) This image depicts a banana and a bowl.}{.201}{.231}%
    {(d) No banana appears in this image.}{.216}{$-$.072}

  \vspace{1pt}\hrule\vspace{1pt}

  \neartierow{}{Negative}%
    {(a) There is no truck in this image.}{.196}{.042}%
    {(c) This image features a truck.}{.211}{$-$.229}

  \vspace{1pt}\hrule\vspace{1pt}

  \neartierow{}{Hybrid}%
    {(a) Shows a frisbee, but no handbag.}{.265}{.150}%
    {(d) No frisbee is present.}{.270}{.006}

  \vspace{1pt}\hrule\vspace{1pt}

  \neartierow{}{Hybrid}%
    {(a) A car is included, but there's no truck.}{.240}{.049}%
    {(c) This image features a truck.}{.258}{$-$.319}

  \caption{\textbf{Near-tie qualitative examples.} Five MCQ questions where CLIP's top-2 scores differ by ${\le}\,0.005$. CLIP selects the wrong answer ({\colorbox{red!12}{red}}); \name{} corrects the ranking ({\colorbox{green!12}{\textbf{green}}} = correct). Two score columns show the ranking flip: CLIP assigns near-identical scores to both options, while our joint model clearly separates them.}
  \label{fig:near_tie_examples}
\end{figure*}

\section{Architecture Details}
\label{sec:architecture}

\Cref{tab:architecture} summarizes the key architectural hyperparameters and parameter counts for both modules.

\begin{table}[h]
  \caption{\textbf{Architecture details} for the ECN and SCN modules (CLIP ViT-B/32, $d{=}512$).}
  \label{tab:architecture}
  \centering
  \footnotesize
  \setlength{\tabcolsep}{4pt}
  \begin{tabular}{@{}l l l@{}}
    \toprule
    & ECN & SCN \\
    \midrule
    Input layers & $l_a{=}6$, $l_p{=}8$ & $\{8,12\}$ \\
    Input features & $\mathbf{h}^{l_p}_{\texttt{[EOS]}}$, $\mathbf{a}$, $\mathbf{h}^{l_a}_{\texttt{[EOS]}}$, $\bar{\mathbf{h}}^{l_p}$ & $\mathbf{f}^{l}_T$, $\mathbf{f}_I$, $\hat{\mathbf{f}}_T$ \\
    Cross-attention & 8 heads, $d{=}512$ & -- \\
    MLP dims & $2048 \to 1024 \to 1024 \to 512$ & $258 \to 128 \to 1$ \\
    Activation & GELU & GELU \\
    Normalization & LayerNorm (post-attn + MLP) & -- \\
    Dropout & 0.1 & -- \\
    Output & $\boldsymbol{\delta} \in \mathbb{R}^{512}$ & $\Delta s \in [-0.2,\, 0.2]$ \\
    Learned scalar & $\alpha$ (init $e^{-0.7} \approx 0.5$) & -- \\
    \#Parameters & 4.7M & 494K \\
    \midrule
    \multicolumn{2}{@{}l}{Total trainable parameters} & 5.2M (3.5\% of CLIP) \\
    \bottomrule
  \end{tabular}
\end{table}

\noindent\textbf{ECN details.}
The cross-attention module uses 8 heads with a single learned query $\mathbf{q} \in \mathbb{R}^{512}$, attending over all $N$ token positions at the peak layer.
A post-attention LayerNorm stabilizes the output before concatenation with the \texttt{[EOS]} and mean-pooled features.
The resulting $4d$-dimensional vector passes through a three-layer bottleneck MLP ($2048 \to 1024 \to 1024 \to 512$) with GELU activations, LayerNorm, and dropout (0.1) after each hidden layer.
The learned scalar $\alpha = \exp(\log\alpha_{\mathrm{init}})$ (initialized to $e^{-0.7} \approx 0.5$) controls the correction magnitude in \cref{eq:ecn_correct}.

\noindent\textbf{SCN details.}
The SCN's 258-dimensional input concatenates two scalar cosine similarities (peak-layer and corrected similarities to the image), a 128-dimensional text context vector encoding the discrepancy between peak and corrected embeddings, and a 128-dimensional cross-modal context vector from the element-wise product of image and corrected text embeddings.
A single hidden layer ($258 \to 128 \to 1$) with GELU activation maps this to a scalar correction bounded by $\tanh$ and $\alpha_{\max}{=}0.20$.

\noindent\textbf{Training hyperparameters.}
Both modules are optimized jointly with AdamW~\cite{loshchilov2019adamw} ($\beta_1{=}0.9$, $\beta_2{=}0.999$, weight decay $0.01$) for 10 epochs.
The ECN uses a learning rate of $1{\times}10^{-5}$ and the SCN uses $1{\times}10^{-4}$ (a 10$\times$ ratio), both with cosine annealing to $1\%$ of the initial rate.
Gradients are clipped at norm $1.0$.
The joint loss weight is $\lambda_{\mathrm{SCN}}{=}1.0$ (\cref{eq:loss_total}), and the InfoNCE logit scale is fixed at $100.0$.
Training uses a batch size of 1024 on a single NVIDIA A100 (40\,GB) and completes in approximately 2.7 hours.

\section{Training Data Construction}
\label{sec:training_data}

Both training data streams are derived from CC12M~\cite{changpinyo2021cc12m} captions using LLaMA~3.1-8B~\cite{grattafiori2024llama} as a negation generator, following the protocol introduced by NegBench~\cite{alhamoud2025negbench}.
CC12M is disjoint from NegBench's evaluation sources (COCO, VOC2007, MSR-VTT): no evaluation image or caption appears in training, and we use NegBench's published generation protocol---identical to the fine-tuning baselines---so any negation-style regularity in the data is shared with the methods we compare against, not specific to \name{}.

\noindent\textbf{Negation generation.}
For each original CC12M caption, we prompt LLaMA~3.1 to produce a semantically negated counterpart by inserting explicit negation cues (``no,'' ``not,'' ``without,'' or restructuring the sentence).
The prompt instructs the model to negate one salient visual attribute or object while keeping the rest of the caption intact, producing captions such as ``a dog sitting on a bench'' $\to$ ``a dog sitting on a bench without a collar.''
We filter out generations that are exact copies, excessively long ($>$77 tokens after CLIP tokenization), or that fail a round-trip consistency check (the negated caption must differ from the original by at least one negation token).

\noindent\textbf{ECN data (contrastive pairs).}
Each original caption and its negated counterpart form a contrastive pair, associated with the original image.
This produces ${\sim}$1.06M training pairs.
During training, both the affirmative and negated captions for each image appear in the same batch, so negated captions serve as hard negatives in the InfoNCE loss.

\noindent\textbf{SCN data (MCQ samples).}
We construct four-option multiple-choice questions by pairing each image with its correct caption and three distractor captions (the negated version plus two randomly sampled captions from other images).
This produces ${\sim}$313K MCQ training samples (78K validation).
The SCN is trained with cross-entropy over the four options plus an $\ell_2$ penalty on $(\Delta s)^2$ to keep corrections small.

\noindent\textbf{Feature pre-extraction.}
To avoid running the frozen CLIP encoder during training, we pre-extract and cache all required features: layer-$L$ \texttt{[EOS]} embeddings, layer-$l_p$ full token sequences, layer-$l_a$ \texttt{[EOS]} embeddings, image embeddings, and projected \texttt{[EOS]} features at each SCN input layer.
This reduces training to lightweight forward passes through the ECN and SCN modules, enabling the full joint training to complete in ${\sim}$2.7 hours on a single NVIDIA A100.

\noindent\textbf{Data statistics.}
After filtering, the final training set contains 1.06M contrastive pairs for the ECN and 313K four-option MCQ samples for the SCN, with a held-out validation split of 78K MCQ samples used for early stopping and hyperparameter selection.
The CC12M source captions cover a broad distribution of visual scenes, objects, and attributes, providing diverse negation patterns that range from simple object absence (``no dog'') to attribute modification (``without the red hat'') and implicit negation (``the empty plate'').
The best checkpoint is selected based on the SCN's validation accuracy on the held-out MCQ split (peak at epoch 8 of 10, 86.65\% 4-way accuracy).

\end{document}